\documentclass[runningheads]{llncs}

\usepackage{graphicx}
\usepackage{comment}
\usepackage{amsmath,amssymb}
\usepackage{color}
\usepackage{url}
\usepackage{hyperref}

\usepackage{graphicx}
\usepackage{booktabs}
\usepackage{subcaption}

\usepackage{ stmaryrd }
\usepackage{textcomp}
\usepackage{gensymb}
\usepackage{multirow}

\newif\ifreview
\reviewfalse

\ifreview
\usepackage{lineno}

\linenumbers
\fi

\begin{document}
	

	\def\SubNumber{000}
	
	\def\GCPRTrack{Special Track: Pattern recognition in the life and natural sciences}

	\title{SealID: Saimaa ringed seal re-identification dataset}

	\ifreview
\else
	
	\author{Ekaterina Nepovinnykh,\inst{1}\orcidID{0000-0002-5045-5041} \and
		Tuomas Eerola\inst{1}\orcidID{0000-0003-1352-0999} \and
		Vincent Biard\inst{2}\orcidID{0000-0003-1885-0946} \and
		Piia Mutka\inst{2} \and
		Marja Niemi\inst{2} \and
		Heikki K\"{a}lvi\"{a}inen\inst{1}\orcidID{0000-0002-0790-6847} \and
		Mervi Kunnasranta\inst{2}}

	\authorrunning{F. Author et al.}
	
	\institute{Computer Vision and Pattern Recognition Laboratory (CVPRL)\\
		Department of Computational and Process Engineering\\
		Lappeenranta-Lahti University of Technology LUT, Lappeenranta, Finland\\
		\email{firstname.lastname@lut.fi}\\ \and
		Department of Environmental and Biological Sciences \\
		University of Eastern Finland \\
		\email{firstname.lastname@uef.fi}}
	\fi
	
	\maketitle              
	
	\begin{abstract}
		Wildlife camera traps and crowd-sourced image material provide novel possibilities to monitor endangered animal species. However, massive image volumes that these methods produce are overwhelming for researchers to go through manually which calls for automatic systems to perform the analysis. The analysis task that has gained the most attention is the re-identification of individuals, as it allows, for example, to study animal migration or to estimate the population size. The Saimaa ringed seal (\textit{Pusa hispida saimensis}) is an endangered subspecies only found in the Lake Saimaa, Finland, and is one of the few existing freshwater seal species. Ringed seals have permanent pelage patterns that are unique to each individual which can be used for the identification of individuals. Large variation in poses further exacerbated by the deformable nature of seals together with varying appearance and low contrast between the ring pattern and the rest of the pelage makes the Saimaa ringed seal re-identification task 
		very challenging, providing a good benchmark to evaluate state-of-the-art re-identification methods. Therefore, we make our Saimaa ringed seal image (SealID) dataset (N=57) publicly available for research purposes. In this paper, the dataset is described, the evaluation protocol for re-identification methods is proposed, and the results for two baseline methods HotSpotter and NORPPA are provided. 
		The SealID dataset has been made publicly available at \url{https://doi.org/10.23729/0f4a3296-3b10-40c8-9ad3-0cf00a5a4a53}.
	\end{abstract}
	
	\section{Introduction}
	\label{sec:intro}
	
	Traditional tools for monitoring animals such as tagging require physical contact with the animal which causes stress, and may change the behavior of the animal. Wildlife photo-identification (Photo-ID) provides tools to study various aspects of animal populations such as migration, survival, dispersal, site fidelity, reproduction, health, population size, or density. The basic idea is to collect image data of a species/population of interest using, for example, digital cameras, game cameras, or crowd-sourcing, to identify the individuals, and to combine the identification with metadata such as the date, the time, and the GPS location of each image. This enables to collect a vast amount of data on populations without using invasive techniques such as tagging. The large scope of the image data calls for automatic solutions, motivating the use of computer vision techniques. From the image analysis point-of-view, the task to be solved is individual re-identification, i.e., finding the matching entry from the database of earlier identified individuals. While human re-identification has been an active research topic for decades, automatic animal re-identification has recently obtained popularity among computer vision researchers.  
	
	The Saimaa ringed seal
	is an endangered species with around 400 individuals alive at the moment \cite{kunnasranta2021sealed}. Due to its conservation status and a small population size, novel monitoring approaches are essential in the development of an effective conservation strategy.
	In the past decade, Photo-ID has been launched as a non-invasive monitoring method for studying population biology and behavior patterns of the Saimaa ringed seal~\cite{koivuniemi2016photo,koivuniemi2019mark}. Ringed seals have a dark pelage ornamented by light grey rings, known as the fur patterns which are permanent and unique to each individual making the re-identification possible.
	
	Saimaa ringed seal image data provide a challenging identification task for developing general-purpose animal re-identification methods that utilize fur, feather, or skin patterns of animals. Large variations in illumination, seal poses, the limited size of identifiable regions, low contrast between the ring pattern and the rest of the pelage, substantial differences between wet and dry fur, and low image quality all contribute to the difficulty of the re-identification task.
	
\begin{figure*}
		\centering
		\begin{subfigure}{0.86\linewidth}
			\includegraphics[width=\linewidth]{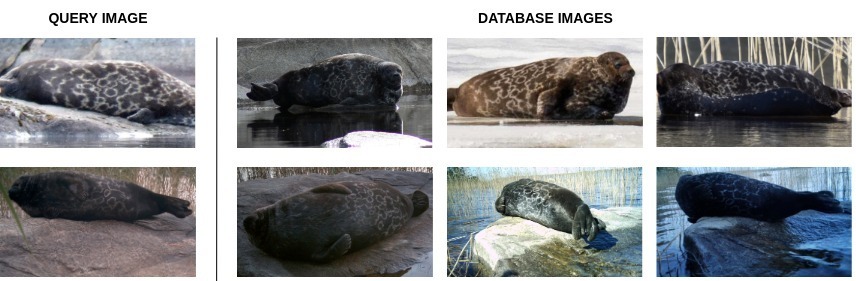}
		\end{subfigure}
		\caption{Examples from the database and the query datasets. Each row contains images of an individual seal. For each image from the query dataset (left), there is the corresponding subset of images from the database (right).}
		\label{fig:dataset_examples}
	\end{figure*}

	We have compiled an extensive dataset of 57 individuals seals, containing a total of 2080 images with individuals identified in each image by an expert and made it publicly available at
	\url{https://doi.org/10.23729/0f4a3296-3b10-40c8-9ad3-0cf00a5a4a53}.
	See Fig.~\ref{fig:dataset_examples} for example images. In this paper, we describe the dataset, propose the evaluation criteria, and present the results for two baseline methods. 

	\section{Related work}
	\subsection{Animal re-identification}
	Camera-based methods utilizing computer vision algorithms have been developed for animal re-identification. Many of them are species-specific which limits their usability~\cite{matthe2017comparison,norouzzadeh2018automatically,giraffe2015}. There have also been research efforts towards creating a unified approach applicable for identification purposes for several animal species. For example, WildMe~\cite{berger2019wildbook,parham2018animal} is a large-scale project for the study, monitoring, and identification of varied species with distinguishable marks on the body. 
	WildMe's re-identification methods are based on the HotSpotter algorithm~\cite{hotspotter}. HotSpotter uses RootSIFT\cite{arandjelovic2012three} descriptors of Affine-invariant regions, spatial reranking with RANSAC, and a scoring mechanism that allows efficient many-to-many matching of images. 
	This algorithm is not species-specific and has been applied to Grevy's and plain zebras (\textit{Equus grevyi}), giraffes (\textit{Giraffa}), leopards (\textit{Panthera pardus}), and lionfish (\textit{Pterois}). 
	
	Due to the recent progress in deep learning, convolutional neural networks (CNN) have become a popular tool for animal biometrics~\cite{schneider2019similarity,schneider2019past}. For example, re-identification of the cattle using CNN combined with k-Nearest Neighbor classifier was proposed in~\cite{bergamini2018multi} where the method was shown to outperform competing methods. The approach is, however, specific to the muzzle patterns of cattle. The muzzle patterns are obtained manually, providing consistent data that simplifies the re-identification. CNN approaches for animal re-identification using natural body markings have been applied to various animals including manta rays~\cite{moskvyak2019robust}, Amur tigers (\textit{Panthera tigris altaica})~\cite{Liu_2019_part,li2019amur,Liu_2019_pose}, zebras (\textit{Equus grevyi}), and giraffes (\textit{Giraffa})~\cite{parham2017animal}. Some species, such as bottlenose dolphins (\emph{Tursiops truncatus}) or African savanna elephants (\emph{Loxodonta africana}) can be identified based on the shape of their body parts, usually their tail or fins, or an ear in case of an elephant. A number of deep learning methods for re-identification are based on this approach: \cite{weideman2020extracting}, CurvRank \cite{weideman2017integral}, finFindR \cite{thompson2019finfindr}, OC/WDTW \cite{bogucki2019applying}.
	
	A typical problem in the wildlife animal re-identification is that it is practically impossible to collect a large dataset with a large number of images for all individuals. Often the method needs to be able to identify an individual with only one or a few previously collected examples. Moreover, the animal re-identification method should be able to recognize if the query image contains an individual that is not in the database of the known individuals.
	Recently, Siamese neural network-based approaches have gained popularity in the animal re-identification~\cite{koch2015siamese}. These methods provide a tool to classify objects based on only one example image (one-shot learning) and to recognize if it belongs to a class that the network has never seen. For example, in~\cite{schneider2019similarity}, the effectiveness of Siamese neural networks for re-identification of humans, chimpanzees, humpback whales, fruit flies, and octopuses was demonstrated. 
	
	\subsection{Saimaa ringed seal re-identification}
	
	A number of studies on the re-identification of ringed seals has been done \cite{zhelezniakov2015segmentation,chehrsimin2017automatic,nepovinnykh2018identification,nepovinnykh2020siamese,chelak2021eden,ladoga,NORPPA}. In~\cite{zhelezniakov2015segmentation}, a superpixel-based segmentation method and a simple texture feature-based ringed seal identification method were presented. 
	In~\cite{chehrsimin2017automatic}, additional preprocessing steps were proposed and two existing species independent individual identification methods were evaluated. However, the identification performance of neither of the methods is good enough for most practical applications. 
	In~\cite{nepovinnykh2018identification}, the re-identification of the Saimaa ringed seals was formulated as a classification problem and was solved using transfer learning. 
	While the performance was high on the used test set, the method is only able to reliably perform the re-identification if there is a large set of examples for each individual. 
	Furthermore, the whole system needs to be retrained if a new seal individual is introduced.
	Finally, it is unclear if the high accuracy was due to the method's ability to learn the necessary features from the fur pattern, or if it also learned features such as pose, size, or illumination that separated individuals in the used dataset, but do not provide the means to generalize the methods to other datasets.
	
	An algorithm for the one-shot re-identification of the Saimaa ringed seals was proposed in~\cite{nepovinnykh2020siamese}. The algorithm consists of the following steps: segmentation, pattern extraction, and patch-based identification. The first step is done using end--to--end semantic segmentation with deep learning. The pattern extraction step relies on 
	the Sato tubeness filter to separate the pattern from the rest of the seal image. The final step is the re-identification. It is done by dividing the pattern into patches and calculating the similarity between them. The patches are compared using a Siamese triplet network. Overall, the system can identify individuals never seen before and shows the promising TOP-5 accuracy, meaning that at least one of the 5 best matches from the database is correct. This algorithm was presented as a part of a larger, species-agnostic re-identification framework. In\cite{chelak2021eden}, a novel pooling layer is proposed to increase the accuracy of patch matching. The idea is to use the eigen decomposition of covariance matrices of features. This method improved the patch and seal re-identification as compared to the previous network architecture in~\cite{nepovinnykh2020siamese}.

	\subsection{Re-identification datasets}
	
	Several 
	publicly available datasets with annotations for animal individuals exist.
	In~\cite{li2019amur}, a novel large-scale Amur tiger re-identification dataset (ATRW) was presented. It contains over 8,000 video clips from 92 individuals with bounding boxes, pose key points and tiger identity annotations. The performance of baseline re-identification algorithms indicates that the dataset is challenging for the re-identification task.

	The 
	ELPephants re-identification dataset\cite{korschens2019elpephants} contains 276 elephant individuals following a long-tailed distribution. It clearly demonstrates challenges for elephant re-identification such as fine-grained differences between the individuals, aging effects on the animals, and large differences in skin color.
	
	In~\cite{beery2019iwildcam}, 
	the 
	iWildCam species identification dataset was described. The dataset consists of nearly 200,000 images collected from various locations and animal species annotated. However, it should be noted that animal individuals are not identified.

	In ~\cite{moskvyak2020learning}, a Manta rays dataset along with a method for the re-identification of Manta rays was proposed. The training set consists of 110 individuals with 1422 images in total. The test set consists of 18 individuals with 321 images in total. The dataset is challenging due to a number of reasons, including large variations in illumination and oblique angles. Those difficulties are similar to the ones encountered in Saimaa ringed seal images. 
	
	To form large and varied datasets, crowdsourcing methods can be used. For example, in~\cite{parham2017animal} the authors propose to use volunteer citizen scientists to collect photos taken in large geographic areas and use computer vision algorithms to semi-automatically identify and count  individual animals. 
	The proposed
	Great Zebra and Giraffe Count and ID dataset contains 4,948 images of only two
	species, Great Zebra (\textit{Equus quagga}) and Masai Giraffe (\textit{Giraffa tippelskirchi}).
	The study in~\cite{borowicz2021social} demonstrated the opportunity of collecting scientifically useful data from the community through the publicly available photo-sharing platform Flickr by creating a dataset for the Weddell seal (\textit{Leptonychotes weddellii}) species.

	\section{Data}
	
	\subsection{Data collection and manual identification}
	
	Data collection was carried out in the Lake Saimaa, Finland (61\degree 05’-62\degree 36’N, 27\degree15’-30\degree 00’ E) under permits by the local environmental authorities (ELY-centre, Metsähallitus). The Photo-ID data were collected annually during the Saimaa ringed seal molting season (mid-April--mid-June) from the year 2010 to 2019 by both ordinary digital cameras (boat surveys) and game camera traps.
	The boat surveys were operated in the main breeding habitat of the Saimaa ringed seals during the first years (Haukivesi since 2010 and Pihlajavesi since 2013) and further covering the whole lake since 2016. Powerboats (a 6--8 m powered boat with a 20--60 hp outboard engine, with one to two observers) were used. A minimum distance of 150 m with the observed seal and the used DSLR cameras (a 55--300 mm telephoto lens) for photographing was kept whenever possible. The GPS coordinates, the observation times, and the numbers of the seals were noted.
	Camera traps were additionally used (Scout Guard SG550, Scout Guard SG560, and Uovision UV785) in Haukivesi (years 2010 to 2012) and Pihlajavesi (since 2013). The game cameras were set in motion sensitivity (2 pictures over a 0.5--2 min time span) or time-laps (2 pictures every 10 min) and were installed in haul-out locations previously found during the boat survey. In case of motion sensitivity cameras, memory cards (2--16GB) were changed 1 to 3 times a week~\cite{koivuniemi2016photo,koivuniemi2019mark}.
	Seal images were matched by an expert using individually characteristic fur patterns.

	\subsection{Data composition}
	\label{DataComposition}
	
	The pelage pattern of the Saimaa ringed seal covers the whole surface area of a seal, making it impossible to see the full pattern from one image. On the other hand, it would be preferable to make a minimal amount of image-to-image comparisons for the re-identification. 
	Ideally, a minimum amount of high-quality images to cover the full view of a seal body is wanted as a set of example images for each known individual.
	
	The dataset is divided into two subsets: the database set and the query set. 
	The database is constructed from the aforementioned minimal sets of high-quality images for each individual seal. Images not included in the database are collected into the query set (N=1650). The query images contain the same individuals as in the database. It was also ensured that query images contain some part of a pattern that could be matched to the visible patterns in the database.
	The database provides a basis for the identification and can be considered as the training set. The query set constitutes the test set used to evaluate the performance of the re-identification methods. Typically, the re-identification algorithm searches for the best match from the database for the given query image.
	
\begin{figure}[ht]
		\centering
		\minipage{0.5\linewidth}
		\includegraphics[width=\linewidth]{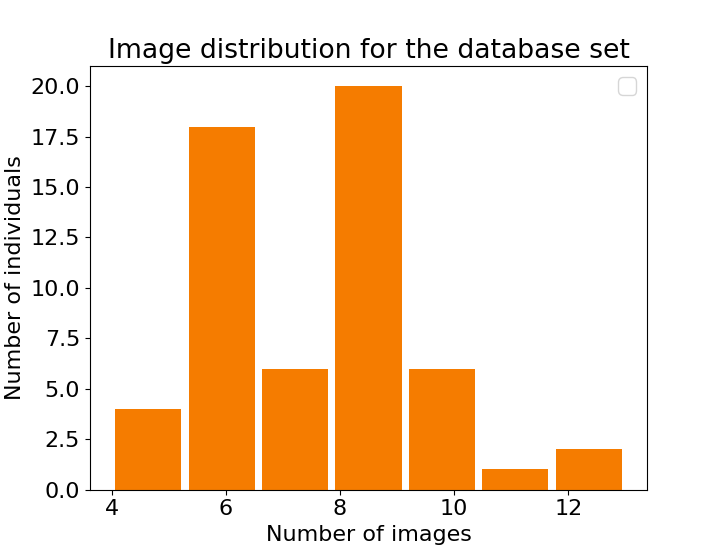}
		\endminipage\hfill
		\minipage{0.5\linewidth}
		\includegraphics[width=\linewidth]{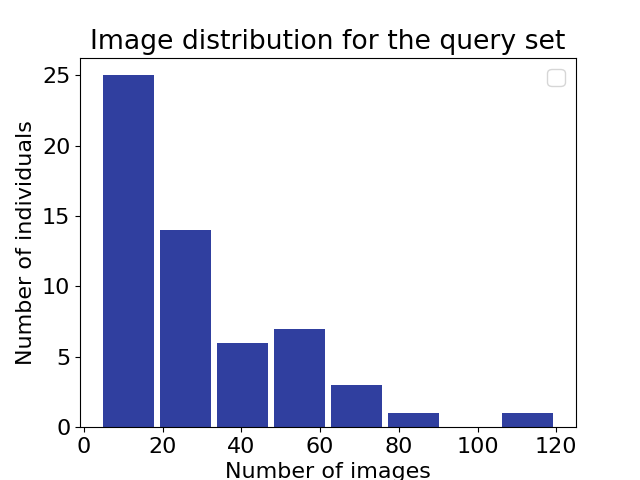}
		\endminipage\hfill
		\caption{Image distributions for the database and query sets.}
		\label{fig:dist}
	\end{figure}

	The total amount of individuals, the total amount of images in the database and in the query sets, the minimum, maximum, mean, and the median number of images per individual for the both sets are presented in Table~\ref{table:image distribution}. Image distributions for the database and query sets is illustrated in Fig.~\ref{fig:dist}.
	Example images are shown in Fig.~\ref{fig:dataset_examples}.

	\begin{table}[ht]
		\centering
		\caption{Image distributions for the database and the query sets.}
		\label{table:image distribution}
		\begin{tabular}{@{}lcc@{}}
			\hline
			\toprule
			& Database set & Query set \\ 
			\midrule
			Total amount of images      & 430          & 1650      \\ 
			Min                         & 4            & 4         \\ 
			Max                         & 13           & 120       \\ 
			Mean                        & 7.5         & 28.9      \\ 
			Median                      & 6            & 42        \\ 
			Total amount of individuals & \multicolumn{2}{c}{57}  \\ 
			\bottomrule
		\end{tabular}
	\end{table}
	
\begin{figure}[ht]
		\centering
		\minipage{0.49\linewidth}
		\includegraphics[width=\linewidth]{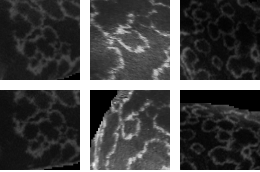}
		\endminipage\hfill
		\minipage{0.49\linewidth}
		\includegraphics[width=\linewidth]{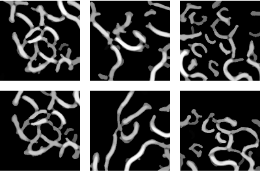}
		\endminipage\hfill
		\caption{Examples of patches: original patches (left); the corresponding pattern patches (right).}
		\label{fig:patches}
	\end{figure}

	A separate dataset with matching patches of the pelage pattern is included to provide the basis for training the pattern matching models. The patch dataset contains, in total, 4599 patches of $60 \times 60$ pixels and is divided into the training and test subsets. The training subset contains 3016 images and 16 classes. The test subset contains 1583 images and 26 classes that are different from the classes in the training set. Each class corresponds to one manually selected location on the pelage pattern and each sample from one class was extracted from different images of the same seal. The Sato tubeness filter based method~\cite{nepovinnykh2020siamese} is applied to each patch to segment the pelage pattern. The extracted pelage pattern patches are manually corrected and included into the dataset. The test set is also divided into the database and query subset with the ratio of 1 to 2. The images that were used to construct the patches dataset are not included into the database and the query subsets of the main re-identification dataset. Examples of patches are presented in Fig.~\ref{fig:patches}.

	\subsection{Seal segmentation}
	
\begin{figure}[ht]
		\centering
		\minipage{0.33\linewidth}
		\includegraphics[width=\linewidth]{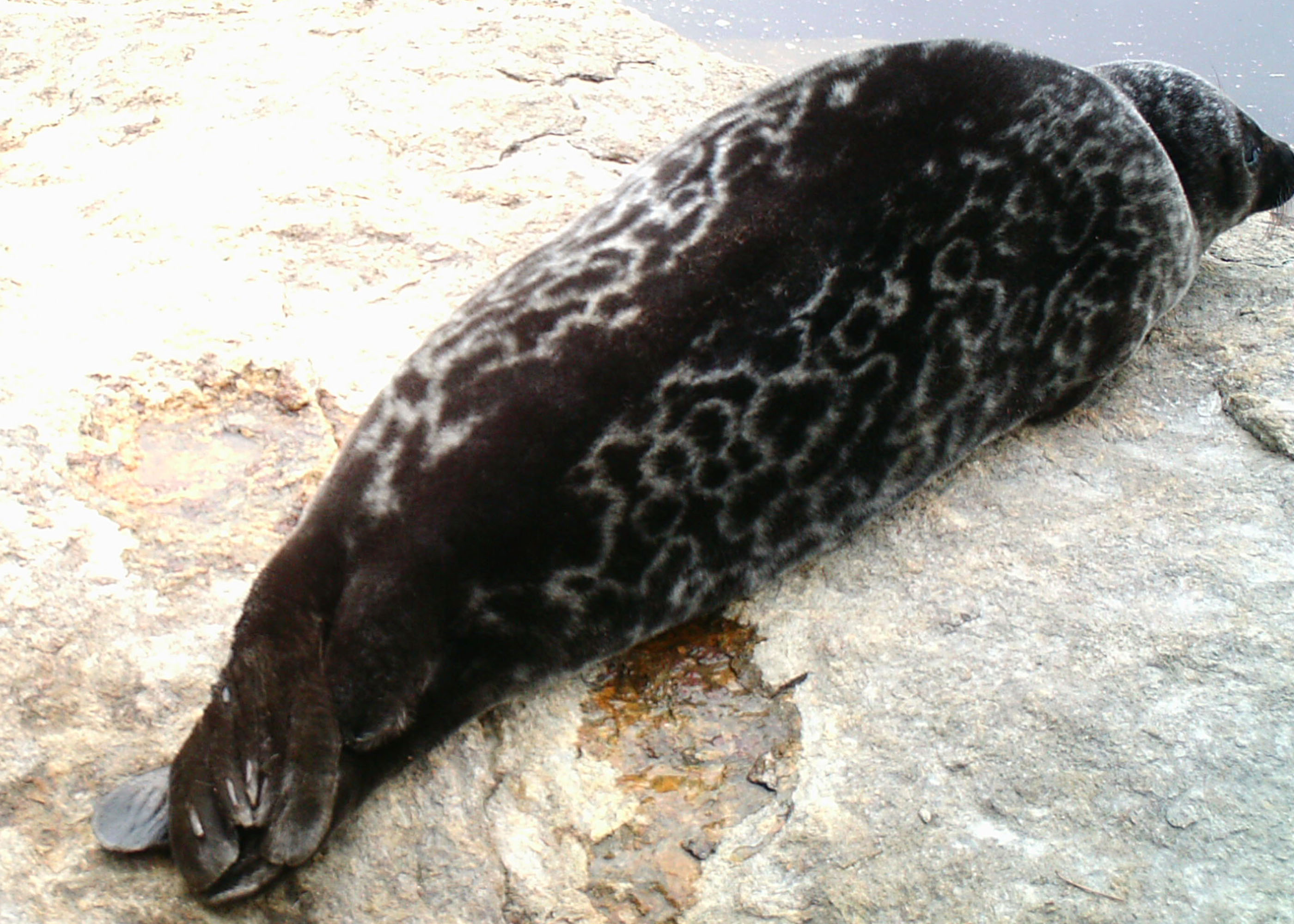}
		\endminipage\hfill
		\minipage{0.33\linewidth}
		\includegraphics[width=\linewidth]{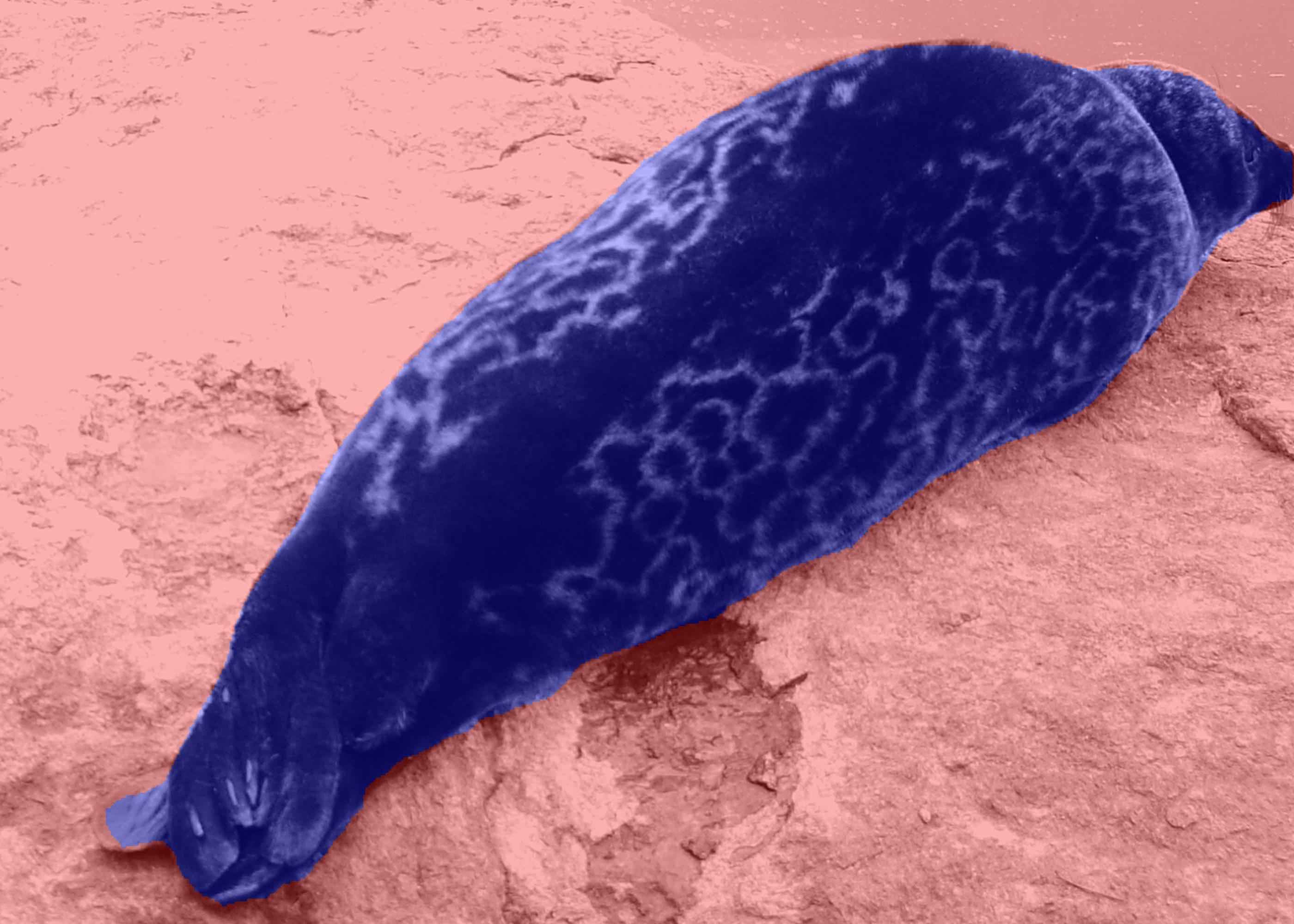}
		\endminipage\hfill
		\minipage{0.33\linewidth}
		\includegraphics[width=\linewidth]{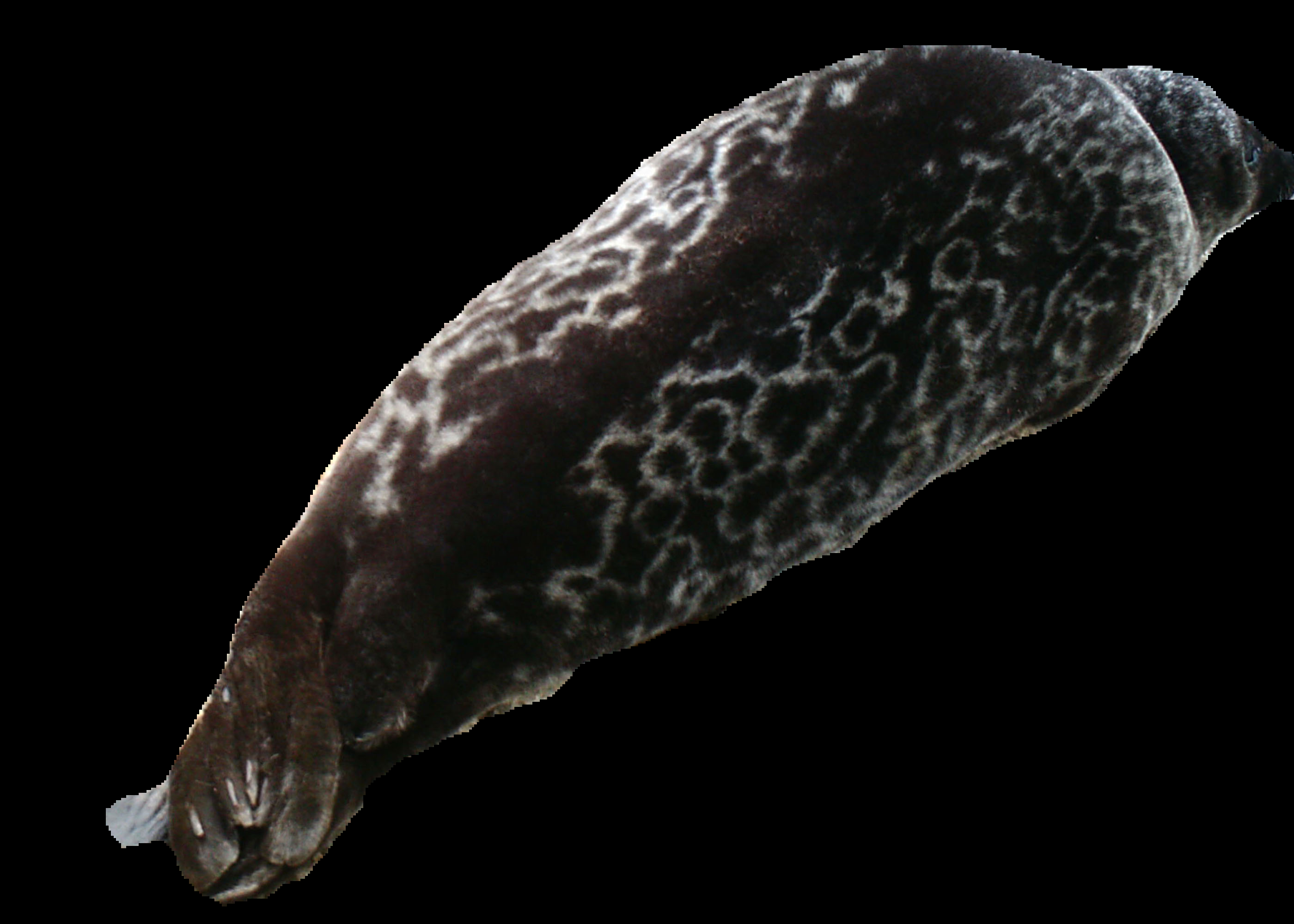}
		
		\endminipage
		\caption{An example of the seal segmentation using the mask: the original image (left), the mask highlighted in blue and the background highlighted in red (middle), and the result of the segmentation (right).}
		\label{fig:segmentation_examples}
	\end{figure}

	The dataset further contains the segmentation masks for each image. The segmentation masks were obtained using a fine-tuned Mask R-CNN model pre-trained on MS COCO dataset~\cite{he2017mask}. The semi-manually segmented datasets of Ladoga and Saimaa ringed seals were used as the ground truth for the transfer learning of the instance seal segmentation~\cite{ladoga}. The results of segmentation were further post-processed in order to fill the holes and to smooth the boundaries of segmented seals, and the segmentation masks were manually corrected. An example of a segmented image from the SealID dataset is presented in Fig.~\ref{fig:segmentation_examples}.
	
	\section{Evaluation protocol}
	
	In this section, an evaluation protocol to enable a fair comparison of methods on the dataset is provided. While the main task is the re-identification, we provide an evaluation protocol also for the segmentation task to allow the benchmarking of segmentation methods using the provided segmentation masks.
	
	\subsection{Segmentation task}
	
\begin{figure}[ht]
		\centering
		\minipage{0.5\linewidth}
		\includegraphics[width=\linewidth]{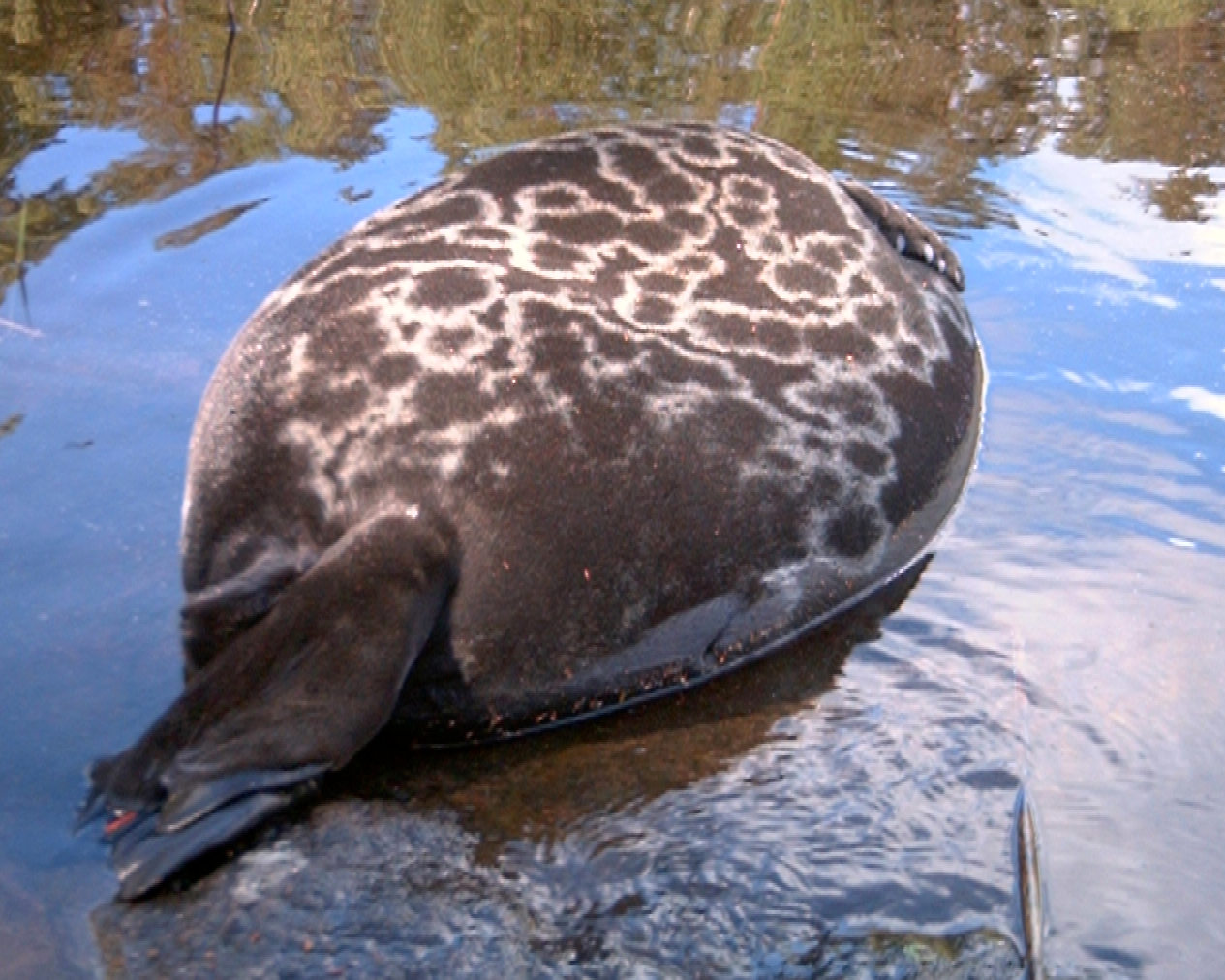}
		\endminipage\hfill
		\minipage{0.5\linewidth}
		\includegraphics[width=\linewidth]{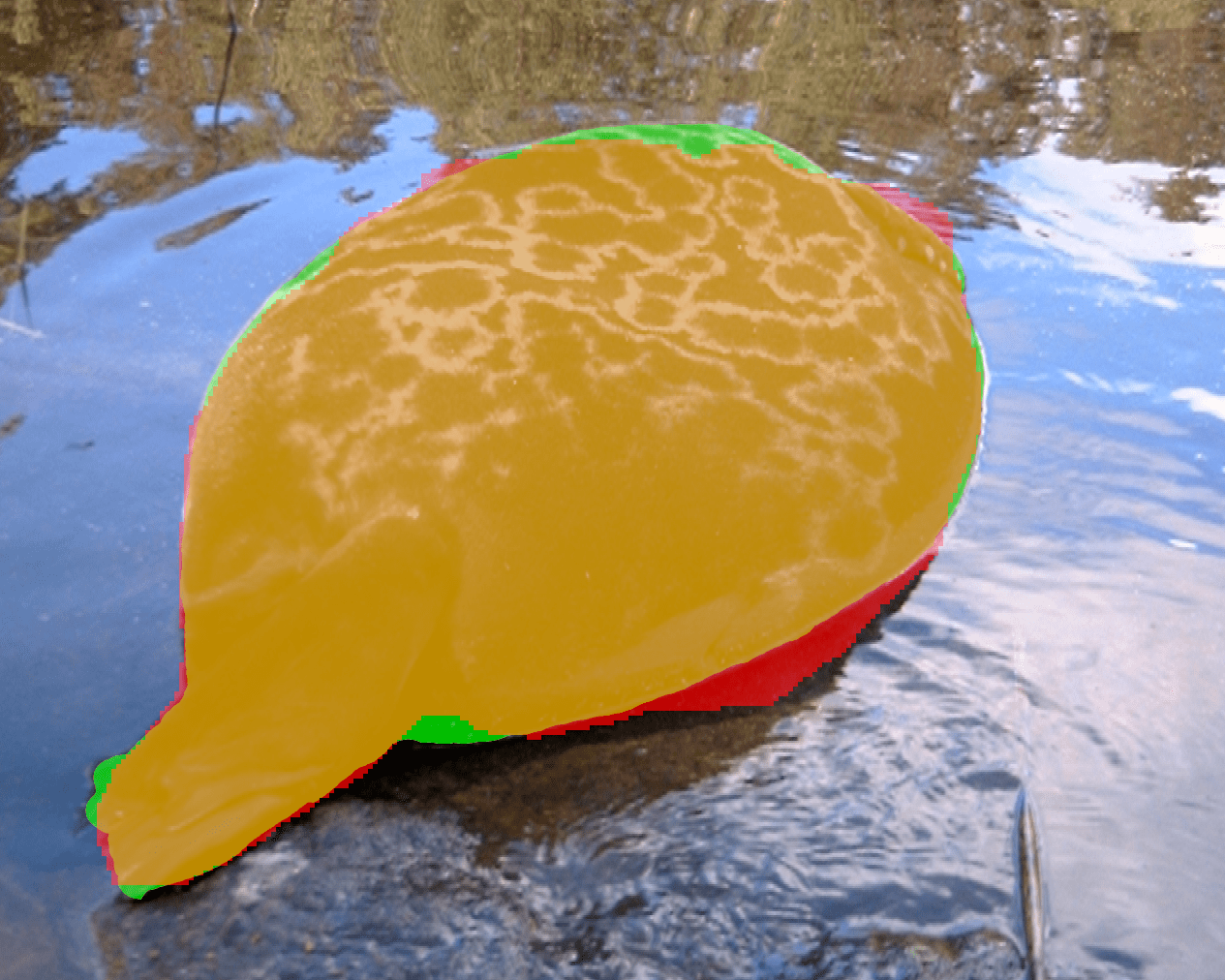}
		\endminipage\hfill
		\caption{Segmentation mask example. The green color depicts ground truth, red is the segmentation result, and yellow is the intersection.}
		\label{fig:iou}
	\end{figure}

	The data for the segmentation task are divided into the training, validation and test sets. In total, 2080 images are used, and the split is 40\% for the training, 20\% for the validation, and 40\% for the testing.
	The performance of the segmentation is evaluated using the Intersection over Union (IoU) metric defined as
	\begin{equation}
	\textrm{IoU}(X,Y) = \frac{|X \cap Y|}{|X \cup Y|}
	\end{equation}
	where $X$ and $Y$ are the sets of pixels from the segmentation result and the ground truth respectively(see Fig.~\ref{fig:iou}).
	
	\subsection{Re-identification task}
	
	Top-$k$ is the primary metric used for the evaluation of the re-identification task. 
	The re-identification is considered correct if any of the $k$ most probable model guesses (best database matches) match to the correct ID. For the model $f$ top-$k$ accuracy is defined as
	\begin{equation}
	\textrm{top-}k = \frac{1}{n} \sum_{x_i\in X} [ y_i \in f_k(x_i) ]
	\end{equation}
	where $X$ is the set of samples, $x_i$ is the $i$-th sample, $n$ is a number of samples,  $y_i$ is its correct label, $f_k(\cdot)$ is the function that returns $k$ most probable guesses for a given sample, and $[\cdot]$ are Iverson brackets, which return 1 if the condition inside is true and 0 otherwise. 
	
	Specifically, top-1, top-3 and top-5 metrics are used. The top-1 accuracy is the conventional accuracy, meaning that the most probable answer is correct. The top-$k$ accuracy can be viewed as a generalization of this. Typically, a re-identification system is deployed in a semi-automatic manner with a biologist verifying the matches. Providing a small set (e.g., 5) of possible matches speeds up the process considerably justifying the use of top-3 and top-5 as an additional evaluation metrics.
	

	\section{Baseline methods}
	
	Two baseline methods were selected. The first one is HotSpotter~\cite{hotspotter} algorithm from WildMe project~\cite{parham2018animal}. It is a unified framework suitable for various species with fur, feather, or skin patterns. The second method is NORPPA~\cite{NORPPA}, a Fisher vector-based pattern matching algorithm which was developed specifically for ringed seals.  
	
	\subsubsection{HotSpotter}
\begin{figure}[ht]
		\centering
		\includegraphics[width=0.9\linewidth]{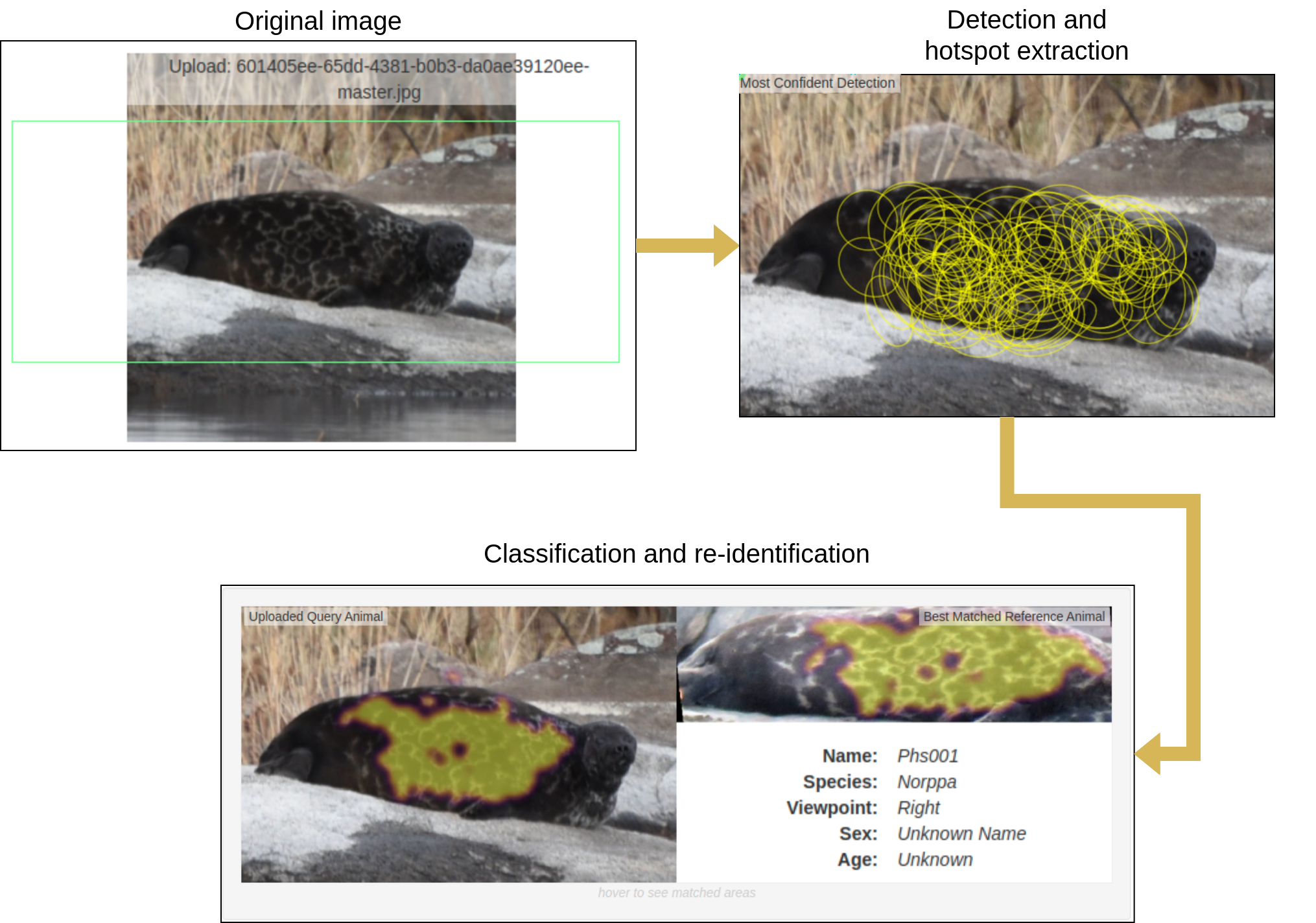}
		\caption{HotSpotter method applied to an image of a Saimaa ringed seal. The matching spots, or "hot spots", are highlighted with yellow. The matching image is chosen according to the score.}
		\label{fig:hotspotter_example}
	\end{figure}

	HotSpotter~\cite{hotspotter} is a SIFT-based~\cite{lowe1999object} algorithm which uses
	viewpoint invariant descriptors and a scoring mechanism that emphasizes the most distinctive keypoints called "hot spots" on an animal pattern. 
	This algorithm has been successfully used for re-identification of zebras (\emph{Equus quagga})~\cite{hotspotter} and giraffes (\emph{Giraffa tippelskirchi})~\cite{parham2017animal}, jaguars (\emph{Panthera onca})~\cite{hotspotter} and ocelots (\emph{Leopardus pardalis})~\cite{nipko2020identifying}. The method is illustrated in Fig.~\ref{fig:hotspotter_example}.

	\subsubsection{Seal re-identification using Fisher Vector (NORPPA)} 
\begin{figure}[ht]
		\centering
		\includegraphics[width=0.8\linewidth]{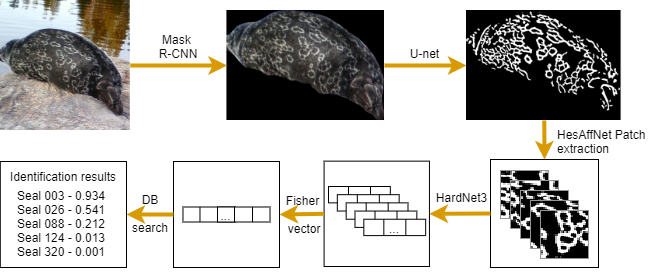}
		\caption{SealID re-identification pipeline~\cite{NORPPA}}
		\label{fig:pipeline}
	\end{figure}

	NOvel Ringed seal re-identification by Pelage Pattern Aggregation (NORPPA) was proposed in~\cite{NORPPA}. The pipeline consists of three main steps: image preprocessing including seal segmentation, extraction of local pelage patterns, and re-identification as shown in Fig.~\ref{fig:pipeline}.. The method utilizes feature aggregation inspired by content based image retrieval techniques~\cite{smeulders2000content}.  HesAffNet~\cite{AffNet2018} patches are embedded using HardNet~\cite{HardNet2017} and aggregated into Fisher vector~\cite{perronnin2007fisher} image descriptors. The final re-identification is performed by calculating cosine distances between Fisher vectors. 
	
	\section{Results}
	
\begin{figure}[ht!]
		\centering
		\includegraphics[width=\linewidth]{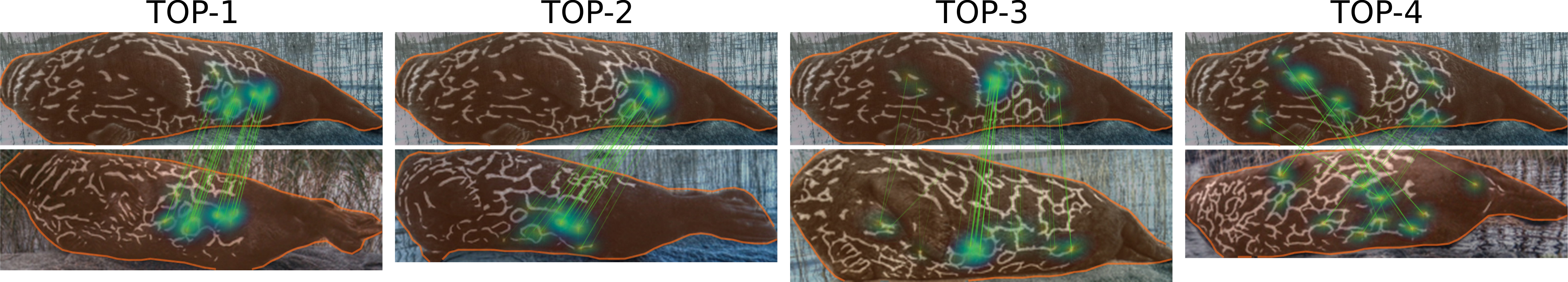}
		\caption{TOP-4 examples for the NORPPA algorithm. First line: query image (phs10), Second line: four best matches in a decreasing order of similarity from left to right. Matched hotspots are highlighted in green. TOP-1--TOP-3 matches are correct. TOP-4 is incorrect.}
		\label{fig:fisher_example}
	\end{figure}

\begin{figure}[ht!]
		\centering
		\includegraphics[width=\linewidth]{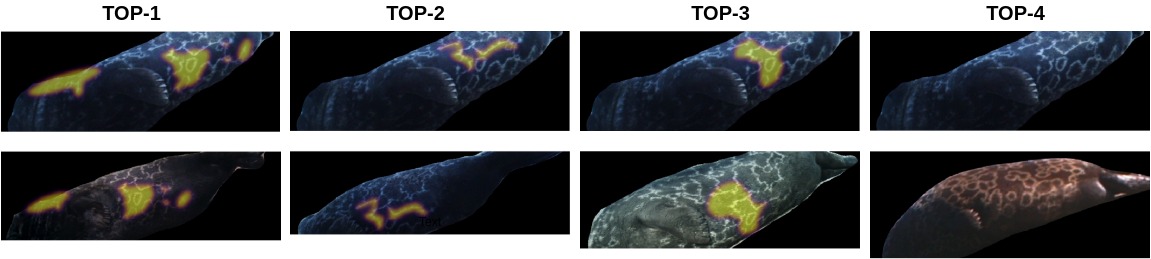}
		\caption{TOP-4 examples for the Hotspotter algorithm. First line: query image (phs10), Second line: four best matches in a decreasing order of similarity from left to right. Matched hotspots are highlighted in yellow. TOP-1--TOP-3 matches are correct. TOP-4 is incorrect.}
		\label{fig:wildbook_example}
	\end{figure}

	The results for the HotSpotter and the NORPPA algorithms are presented in Table~\ref{table:reid}. For both approaches the experiments were executed with and without preprocessing step which consists of tone mapping and segmentation as described in~\cite{NORPPA}. It is clear that preprocessing improves the results for HotSpotter, but it is even more important for NORPPA. Even though the accuracy of NORPPA without the preprocessing step is much lower than the accuracy of HotSpotter, with the preprocessing step NORPPA outperforms HotSpotter by a notable margin. 
	Examples of the results of the NORPPA and HotSpotter algorithms are presented in Fig.~\ref{fig:fisher_example} and  Fig.~\ref{fig:wildbook_example} respectively.

	\begin{table}[ht]
		\centering
		\caption{Re-identification results with HotSpotter and NORPPA.}
		\label{table:reid}
		\begin{tabular}{|l|l|c|c|c|}
			\hline
			&                                           & TOP-1 \quad   & TOP-3  \quad  & TOP-5 \quad  \\\hline
			\multirow{2}{*}{HotSpotter} & Raw           & 61.87         & 63.63         & 64.42          \\ \cline{2-5} & Preprocessed                  & 69.39         & 72.00         & 73.15          \\ \hline
			\multirow{2}{*}{NORPPA} & Raw               & 49.52         & 59.58         & 64.55          \\ \cline{2-5} 
			& Preprocessed      & \textbf{77.64\%} & \textbf{82.97\%} & \textbf{85.27\%} \\ \hline
		\end{tabular}
	\end{table}
	
	\section{Conclusion}
	
	In this paper, the Saimaa ringed seal re-identification dataset (SealID) was presented. Compared to other published animal re-identification datasets, the SealID dataset provides a more challenging identification task due to the large variation in illumination, seal poses, the limited size of identifiable regions, low contrast between the ring pattern, and substantial variations in the appearance of the pattern. Therefore, the database allows to push forward the development of general-purpose animal re-identification methods for wildlife conservation. The dataset contains a curated gallery database with example images of each seal individual and a large set of challenging query images to be re-identified. The segmentation masks are provided for both database and gallery images. A separate dataset of pelage pattern patches is included in the database. We further propose the evaluation protocol to allow a fair comparison between methods and show results for two baseline methods HotSpotter and NORPPA. The results demonstrate the challenging nature of the data, but also show the potential of modern computer vision techniques in the re-identification task. We have made the database publicly available for other researchers.
	
	
	

	%
	%
	%
	%
	\bibliographystyle{splncs04}
	\bibliography{main}
	
\end{document}